\documentclass[11pt,a4paper]{article}
\usepackage[hyperref]{emnlp2020}
\usepackage{times}
\usepackage{latexsym}

\usepackage{graphicx}
\usepackage{graphics}
\usepackage{microtype}
\usepackage{booktabs}
\usepackage{amsmath}
\usepackage{enumitem}
\usepackage{multirow}
\usepackage{multicol}
\usepackage{xspace}
\newcommand{\nop}[1]{}

\newcommand{\dataset}{\textsc{HybridQA}\xspace}
\newcommand{\model}{\textsc{hybrider}\xspace}

\aclfinalcopy 

\title{HybridQA: A Dataset of Multi-Hop Question Answering \\over Tabular and Textual Data}

\author{Wenhu Chen, Hanwen Zha, Zhiyu Chen, Wenhan Xiong, Hong Wang, William Wang\\
University of California, Santa Barbara, CA, USA\\
\tt{\{wenhuchen, hwzha, xwhan\}@cs.ucsb.edu,}\\
\tt{\{zhiyuchen, hongwang600, william\}@cs.ucsb.edu}\\
}

\date{}

\begin{document}
\maketitle
\begin{abstract}
Existing question answering datasets focus on dealing with homogeneous information, based either only on text or KB/Table information alone. However, as human knowledge is distributed over heterogeneous forms, using homogeneous information alone might lead to severe coverage problems. To fill in the gap, we present HybridQA\footnote{\url{https://github.com/wenhuchen/HybridQA}}, a new large-scale question-answering dataset that requires reasoning on heterogeneous information. Each question is aligned with a Wikipedia table and multiple free-form corpora linked with the entities in the table. The questions are designed to aggregate both tabular information and text information, i.e., lack of either form would render the question unanswerable. We test with three different models: 1) a table-only model. 2) text-only model. 3) a hybrid model that combines heterogeneous information to find the answer. The experimental results show that the EM scores obtained by two baselines are below 20\%, while the hybrid model can achieve an EM over 40\%. This gap suggests the necessity to aggregate heterogeneous information in HybridQA. However, the hybrid model's score is still far behind human performance. Hence, HybridQA can serve as a challenging benchmark to study question answering with heterogeneous information. 
\end{abstract}

\section{Introduction}
Question answering systems aim to answer any form of question of our interests, with evidence provided by either free-form text like Wikipedia passages~\cite{rajpurkar2016squad,chen2017reading,yang2018hotpotqa} or structured data like Freebase/WikiData~\cite{berant2013semantic,kwiatkowski2013scaling,yih2015semantic,weston2015towards} and WikiTables~\cite{pasupat2015compositional}. Both forms have their advantages, the free-form corpus has in general better coverage while structured data has better compositionality to handle complex multi-hop questions. Due to the advantages of different representation forms, people like to combine them in real world applications. Therefore, it is sometime not ideal to assume the question has answer in a passage. This paper aims to simulate a more realistic setting where the evidences are distributed into heterogeneous data, and the model requires to aggregate  information from different forms for answering a question. There has been some pioneering work on building hybrid QA systems~\cite{sun2019pullnet,sun2018open,xiong2019improving}. These methods adopts KB-only datasets~\cite{berant2013semantic,yih2015semantic,talmor2018web} to simulate a hybrid setting by randomly masking KB triples and replace them with text corpus. Experimental results have proved decent improvement, which shed lights on the potential of hybrid question answering systems to integrate heterogeneous information.
\begin{table}[!t]
\small
\begin{tabular}{lcccc}
\toprule
Dataset      & Size & \#Docs       & \begin{tabular}[c]{@{}l@{}}KB/\\ Table\end{tabular} & Multi-Hop \\
\midrule
WebQuestions & 5.8K   & no              & yes                                                 & yes       \\
WebQSP       & 4.7K   & no              & yes                                                 & yes       \\
WebQComplex  & 34K    & no              & yes                                                 & yes       \\
MetaQA       & 400k   & no              & yes                                                 & yes       \\
WikiTableQA  & 22K    & no              & yes                                                 & yes       \\
\midrule
SQuAD-v1     & 107K   & 1               & no                                                  & no        \\
DROP         & 99K    & 1               & no                                                  & yes       \\
TriviaQA     & 95K    & \textgreater{}1 & no                                                  & no        \\
HotpotQA    & 112K   & \textgreater{}1 & no                                                  & yes       \\
Natural-QA   & 300K   & \textgreater{}1 & no                                                  & yes        \\
\midrule
\dataset     & 70K    & $>>$1 & yes                                                 & yes     \\     
\bottomrule
\end{tabular}
\caption{Comparison of existing datasets, where \#docs means the number of documents provided for a specific question. 1) KB-only datasets: WebQuestions~\cite{berant2013semantic}, WebQSP~\cite{yih2016value}, WebComplex~\cite{talmor2018web}, MetaQA~\cite{zhang2018variational}, WikiTableQuestion~\cite{pasupat2015compositional}. 2) Text-only datasets with single passage: like SQuAD~\cite{rajpurkar2016squad}, DROP~\cite{dua2019drop}. 3) open-domain Text-Only dataset: TriviaQA~\cite{joshi2017triviaqa}, HotpotQA~\cite{yang2018hotpotqa}, Natural Questions~\cite{kwiatkowski2019natural}.}
\label{tab:dataset}
\vspace{-3ex}
\end{table}

\begin{figure*}[!t]
    \centering
    \includegraphics[width=1.0\linewidth]{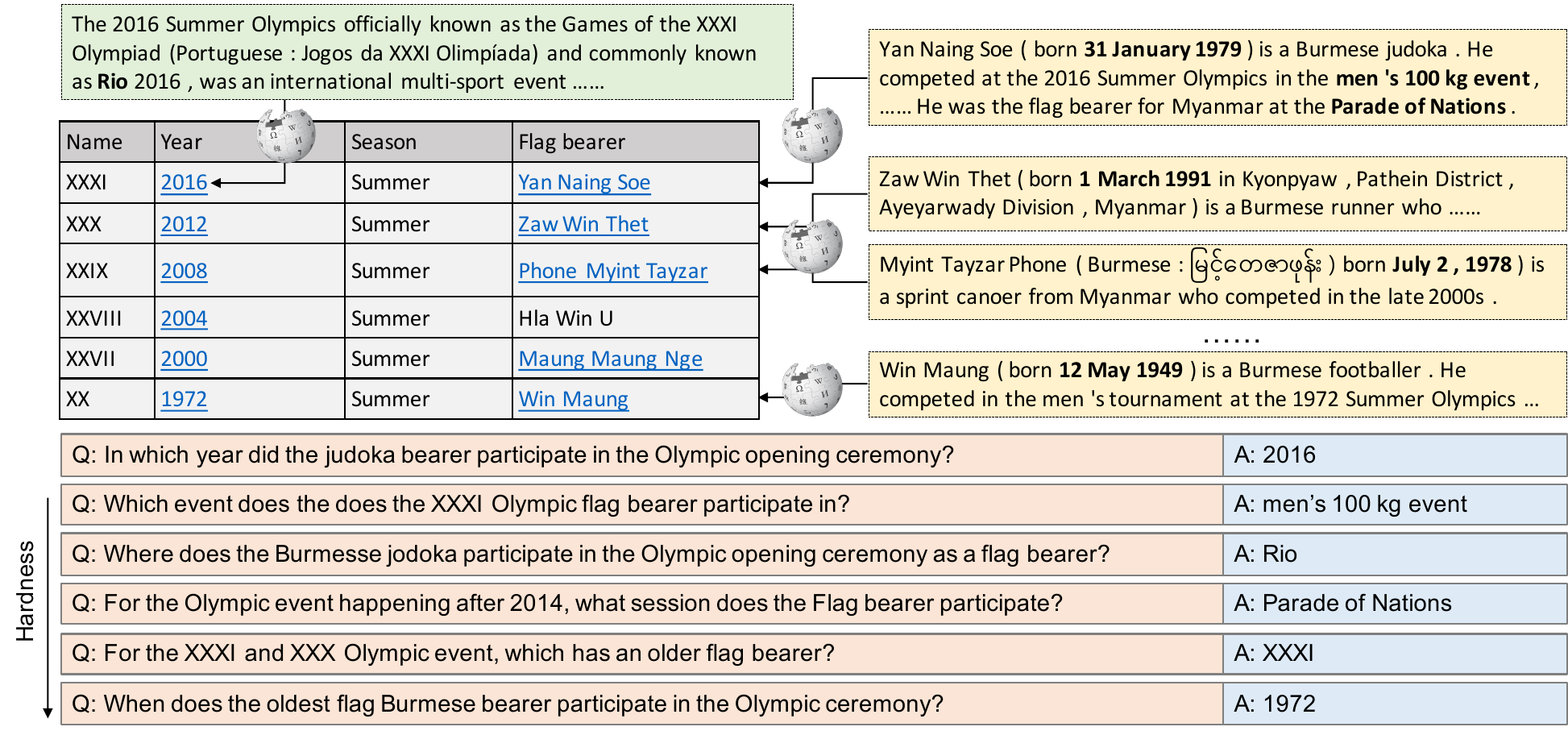}
    \caption{Examples of annotated question answering pairs from Wikipedia page\footnote{\url{https://en.wikipedia.org/wiki/List_of_flag_bearers_for_Myanmar_at_the_Olympics}}. Underlined entities have hyperlinked passages, which are displayed in the boxes. The lower part shows the human-annotated question-answer pairs roughly categorized based on their hardness.}
    \label{fig:example}
    \vspace{-2ex}
\end{figure*}
Though there already exist numerous valuable questions answering datasets as listed in~\autoref{tab:dataset}, these datasets were initially designed to use either structured or unstructured information during annotation. There is no guarantee that these questions need to aggregate heterogeneous information to find the answer. Therefore, designing hybrid question answering systems would probably yield marginal benefits over the non-hybrid ones, which greatly hinders the research development in building hybrid question answering systems.

To fill in the gap, we construct a heterogeneous QA dataset \dataset, which is collected by crowdsourcing based on Wikipedia tables. During annotation, each crowd worker is presented with a table along with its hyperlinked Wikipedia passages to propose questions requiring aggregating both forms of information. The dataset consists of roughly 70K question-answering pairs aligned with 13,000 Wikipedia tables. As Wikitables~\cite{bhagavatula2013methods} are curated from high-standard professionals to organize a set of information regarding a given theme, its information is mostly absent in the text. Such a complementary nature makes WikiTables an ideal environment for hybrid question answering. To ensure that the answers cannot be hacked by single-hop or homogeneous models, we carefully employ different strategies to calibrate the annotation process. An example is demonstrated in~\autoref{fig:example}. This table is aimed to describe Burmese flag bearers over different Olympic events, where the second column has hyperlinked passages about the Olympic event, and the fourth column has hyperlinked passages about biography individual bearers. The dataset is both multi-hop and hybrid in the following senses: 1) the question requires multiple hops to achieve the answer, each reasoning hop may utilize either tabular or textual information. 2) the answer may come from either the table or a passage. 

In our experiments, we implement three models, namely Table-only model, Passage-only, and a heterogeneous model \model, which combines both information forms to perform multi-hop reasoning. Our Experiments show that two homogeneous models only achieve EM lower than 20\%, while \model can achieve an EM over 40\%, which concludes the necessity to do multi-hop reasoning over heterogeneous information on \dataset. As the \model is still far behind human performance, we believe that it would be a challenging next-problem for the community.

\begin{table}[!t]
\small
\centering
\begin{tabular}{lccc}
\hline
\#Table & \#Row/\#Column & \#Cell &  \#Links/Table \\ 
\hline
13,000 & 15.7/4.4 & 70 & 44 \\
\hline
\#Passage & \#Words/Passage & \#Ques & \#Words/Ques \\
\hline
293,269 & 103 &  69,611 & 18.9 \\
\hline
\end{tabular}
\caption{Statistics of Table and Passage in our dataset.}
\label{tab:table_and_passage}
\vspace{-2ex}
\end{table}

\section{Dataset}
In this section, we describe how we crawl high-quality tables with their associated passages, and then describe how we collect hybrid questions. The statistics of \dataset is in~\autoref{tab:table_and_passage}.

\paragraph{Table/Passage Collection}
To ease the annotation process, we apply the following rules during table crawling. 1) we need tables with rows between 5-20, columns between 3-6, which is appropriate for the crowd-workers to view. 2) we restrain the tables from having hyperlinked cells over 35\% of its total cells, which provide an abundant amount of textual information. For each hyperlink in the table, we retrieve its Wikipedia page and crop at most the first 12 sentences from its introduction session as the associated passage. 3) we apply some additional rules to avoid improper tables and finally collect 13,000 high-quality tables.  

\paragraph{Question/Answer Collection}
We release 13K HITs (human intelligence task) on the Amazon Mechanical Turk platform, where each HIT presents the crowd-worker with one crawled Wikipedia table along with its hyperlinked passages. We require the worker to write six questions as well as their answers. The question annotation phase is not trivial, as we specifically need questions that rely on both tabular and textual information. In order to achieve that, we exemplify abundant examples in our Amazon Turker interface with detailed explanations to help crowd-workers to understand the essence of the ``hybrid" question. The guidelines are described as follows:
\begin{itemize}
    \item The question requires multiple steps over two information forms of reasoning to answer.
    \item Table reasoning step specifically includes (i) filter our table rows based on equal/greater/less, e.g. ``For the XXXI Olympic event", (ii)) superlative operation over a column, e.g. ``the earliest Olympic event", (iii) hop between two cells, e.g. ``Which event ... participate in ...", (iv) extract information from table, e.g. ``In which year did the player ... ".
    \item Text reasoning step specifically includes (i) select passages based on the certain mentions, e.g. ``the judoka bearer", (ii) extract a span from the passage as the answer.
    \item The answer should be a minimum text span from either a table cell or a specific passage.
\end{itemize}
Based on the above criteria, we hire five CS-majored graduate students as our ``human expert" to decide the acceptance of a HIT. The average completion time for one HIT is 12 minutes, and payment is \$2.3 U.S. dollars/HIT. 

\paragraph{Annotation De-biasing}
As has been suggested in previous papers~\cite{kaushik2018much,chen2019understanding,clark2019don}, the existing benchmarks on multi-hop reasoning question answering have annotation biases, which makes designing multi-hop models unnecessary. We discuss different biases and our prevention as follows:
\begin{itemize}
    \item \textit{Table Bias}: our preliminary study observes that the annotators prefer to ask questions regarding the top part of the table. In order to deal with this issue, we explicitly highlight certain regions in the table to encourage crowd-workers to raise questions regarding the given uniformly-distributed regions.
    \item \textit{Passage Bias}: the preliminary study shows that the annotators like to ask questions regarding the first few sentences in the passage. In order to deal with such a bias, we use an algorithm to match the answer with linked passages to find their span and reject the HITs, which have all the answers centered around the first few sentences.
    \item \textit{Question Bias}: the most difficult bias to deal with is the ``fake" hybrid question like ``when is 2012 Olympic Burmese runner flag bearer born?" for the table listed in~\autoref{fig:example}. Though it seems that ``2012 Olympic" is needed to perform hop operation on the table, the ``runner flag bearer" already reviews the bearer as ``Zaw Win Thet" because there is no other runner bearer in the table. With that said, reading the passage of ``Zaw Win Thet" alone can simply lead to the answer. In order to cope with such a bias, we ask ``human experts" to spot such questions and reject them.
\end{itemize}

\paragraph{Statistics}
After we harvest the human annotations from 13K HITs (78K questions), we trace back the answers to its source (table or passage). Then we apply several rules to further filter out low-quality annotations: 1) the answer cannot be found from either table or passage, 2) the answer is longer than 20 words, 3) using a TF-IDF retriever can directly find the answer passage with high similarity without relying on tabular information. 

We filter the question-answer pairs based on the previous criteria and release the filtered version. As our goal is to solve multi-hop hybrid questions requiring a deeper understanding of heterogeneous information. We follow HotpotQA~\cite{yang2018hotpotqa} to construct a more challenging dev/test split in our benchmark. Specifically, we use some statistical features like the ``size of the table", ``similarity between answer passage and question", ``whether question directly mentions the field", etc. to roughly classify the question into two difficulty levels: simple (65\%) and hard (35\%). We construct our dev and test set by sampling half-half from the two categories. We match the answer span against all the cells and passages in the table and divide the answer source into three categories: 1) the answer comes from a text span in a table cell, 2) the answer comes from a certain linked passage, 3) the answer is computed by using numerical operation like `count', `add', `average', etc. The matching process is approximated, not guaranteed to be 100\% correct. We summarize our findings in~\autoref{tab:data_split}. In the following experiments, we will report the EM/F1 score for these fine-grained question types to better understand our results.

\begin{table}[!ht]
\small
\centering
\begin{tabular}{lccccc}
\toprule
Split & Train & Dev & Test & Total \\
\midrule
In-Passage  &  35,215 & 2,025  & 20,45  & 39,285 (56.4\%) \\
In-Table & 26,803 & 1,349 & 1,346 & 29,498 (42.3\%)  \\
Computed & 664  & 92 & 72 & 828 (1.1\%) \\
Total & 62,682 & 3,466 & 3,463 & 69,611 \\
\bottomrule
\end{tabular}
\caption{Data Split: In-Table means the answer comes from plain text in the table, and In-Passage means the answer comes from certain passage. }
\label{tab:data_split}
\vspace{-2ex}
\end{table}

\section{Data Analysis}
In this section, we specifically analyze the different aspects of the dataset to provide the overall characteristics of the new dataset.
\subsection{Question Types}
\begin{figure}[thb]
    \centering
    \includegraphics[width=1.0\linewidth]{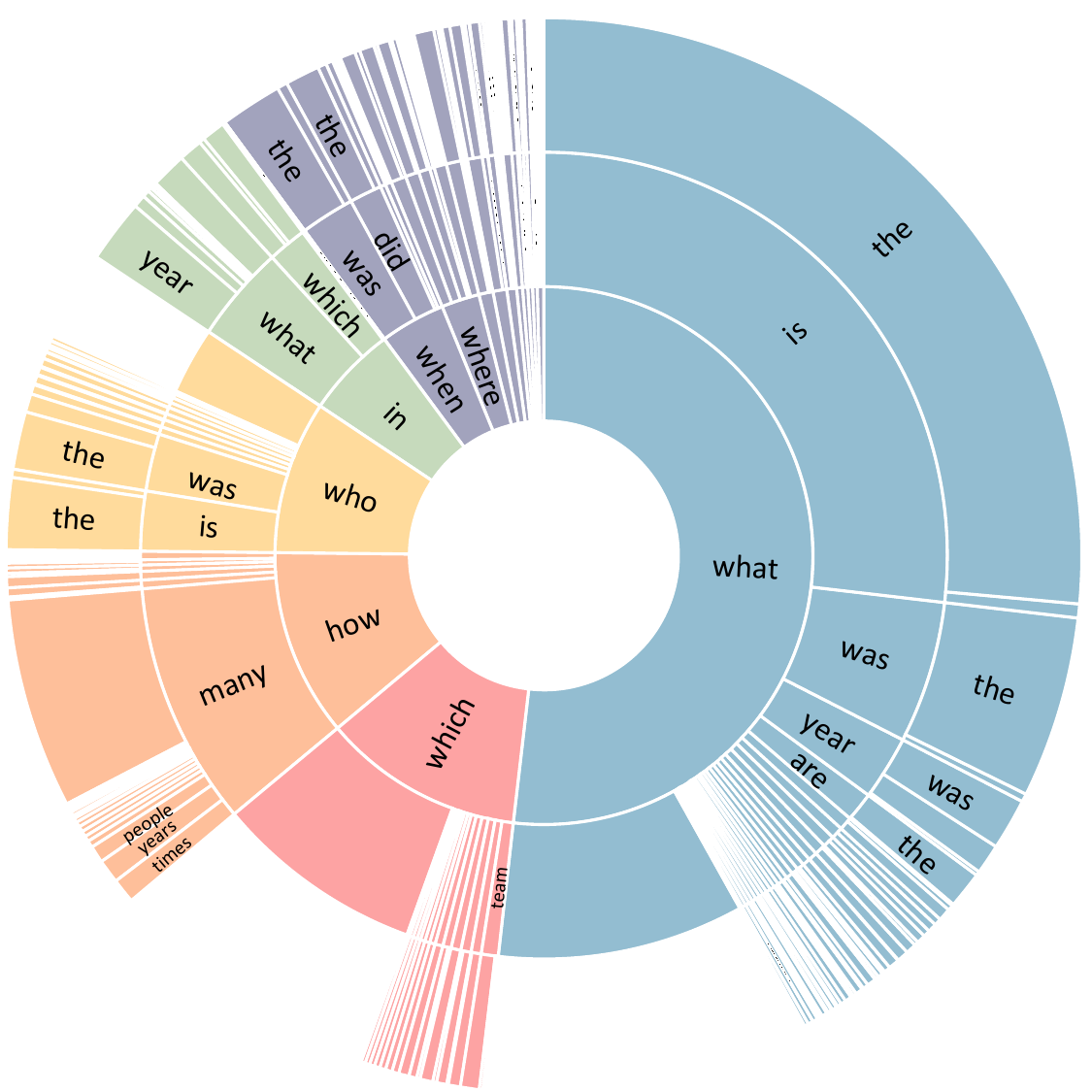}
    \caption{The type of questions in \dataset, question types are extracted using rules starting at the question words or preposition before them. }
    \label{fig:question_type}
\end{figure}

We heuristically identified question types for each collected question. To identify the question type, we locate the central question word (CQW) in the question and take the neighboring three tokens~\cite{yang2018hotpotqa} to determine the question types. We visualize the distribution in~\autoref{fig:question_type}, which demonstrates the syntactic diversity of the questions in \dataset.

\subsection{Answer Types}
We further sample 100 examples from the dataset and present the types of answers in~\autoref{tab:answer-types}. As can be seen, it covers a wide range of answer types. Compared to~\cite{yang2018hotpotqa}, our dataset covers more number-related or date-related questions, which reflects the nature of tabular data.

\begin{table}[htbp]
\small
\begin{center}
\begin{tabular}{lccc}
\toprule
Answer Type & \% & Example(s) \\
\midrule
Location  & 22  & Balestier Road, Atlanta\\
Number  & 22  & 762 metres ( 2,500 ft ), 15,000\\
Date  & 20  & April 25 , 1994, 1913\\
Person  & 15  & Bärbel Wöckel, Jerry\\
Group  & 3  & Hallmark Entertainment\\
Event  & 3  & Battle of Hlobane\\
Artwork  & 1  & Songmaster\\
Adjective & 4 & second-busiest\\
Other proper noun  & 8  & Space Opera, CR 131\\
Common noun  & 1  & other musicians\\
\bottomrule
\end{tabular}
\end{center}
\caption{Types of answers in \textsc{HyrbidQA}.}
\label{tab:answer-types}
\end{table}

\subsection{Inference Types}
\begin{figure*}[thb]
    \centering
    \includegraphics[width=1.0\linewidth]{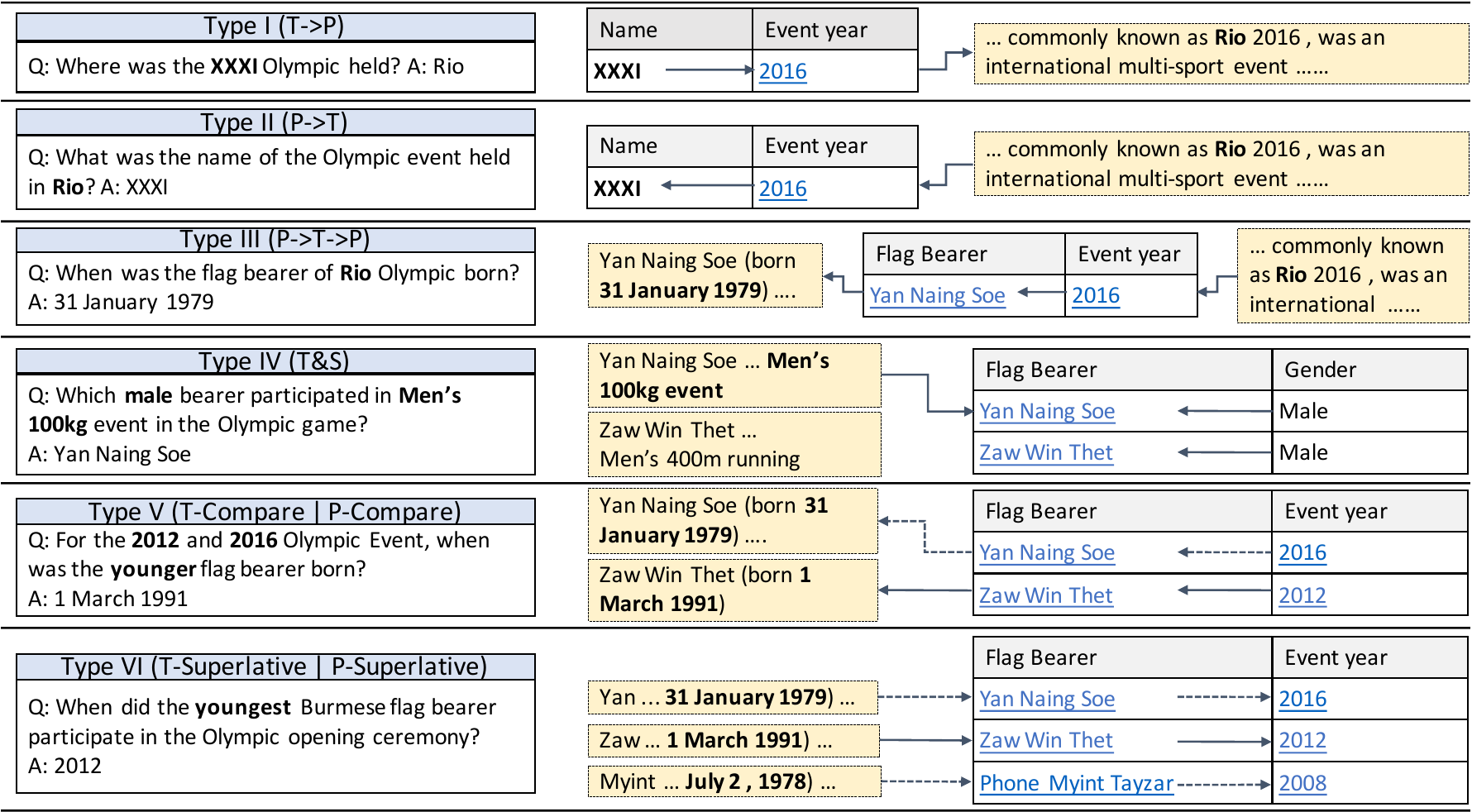}
    \caption{Illustration of different types of multi-hop questions.}
    \label{fig:types}
    \vspace{-2ex}
\end{figure*}

We analyze multi-hop reasoning types in~\autoref{fig:types}. According to our statistics, most of the questions require two or three hops to find the answer.

\noindent 1) Type I question (23.4\%)  uses Table $\rightarrow$ Passage chain, it first uses table-wise operations (equal/greater/less/first/last/argmax/argmin) to locate certain cells in the table, and then hop to their neighboring hyperlinked cells within the same row, finally extracts a text span from the passage of the hyperlinked cell as the answer. \\
\noindent 2) Type II question (20.3\%) uses Passage $\rightarrow$ Table chain, it first uses cues present in the question to retrieve related passage, which traces back to certain hyperlinked cells in the table, and then hop to a neighboring cell within the same row, finally extracts text span from that cell. \\
\noindent 3) Type III question (35.1\%) uses Passage $\rightarrow $Table$ \rightarrow $Passage chain, it follows the same pattern as Type II, but in the last step, it hops to a hyperlinked cell and extracts answer from its linked passage. This is the most common pattern. \\
\noindent 4) Type IV question (17.3\%) uses Passage and Table jointly to identify a hyperlinked cell based on table operations and passage similarity and then extract the plain text from that cell as the answer. \\
\noindent  5) Type V question (3.1\%) involves two parallel reasoning chain, while the comparison is involved in the intermediate step to find the answer. \\
\noindent  6) Type VI questions (0.8\%) involve multiple reasoning chains, while superlative in involved in the intermediate step to obtain the correct answer.

\section{Model}
In this section, we propose three models we use to perform question answering on \dataset.
\subsection{Table-Only Model}
\begin{figure}[!t]
    \centering
    \includegraphics[width=1.0\linewidth]{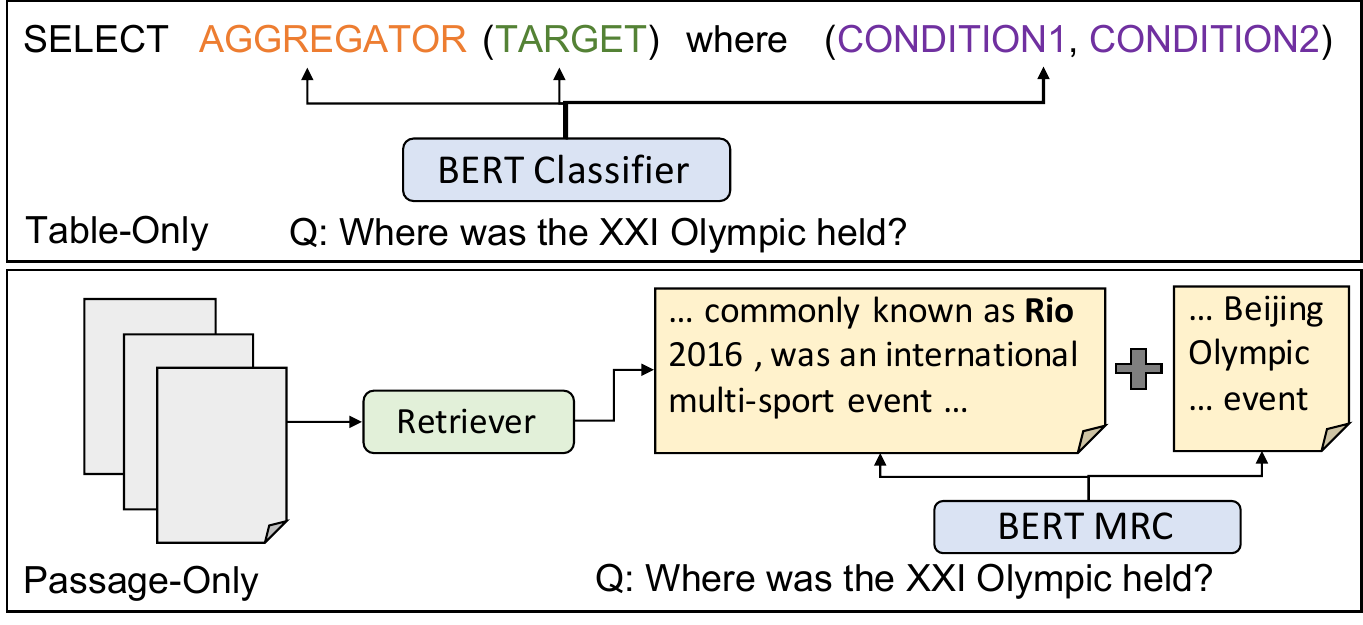}
    \caption{Illustration of the table-only and passage-only baselines, both are based on BERT Encoder.}
    \label{fig:baseline}
    \vspace{-2ex}
\end{figure}

In this setting, we design a model that can only rely on the tabular information to find the answer. Our model is based on the SQL semantic parser~\cite{zhong2017seq2sql,xu2017sqlnet}, which uses a neural network to parse the given questions into a symbolic form and execute against the table. We follow the SQLNet~\cite{xu2017sqlnet} to flatten the prediction of the whole SQL query into a slot filling procedure. More specifically, our parser model first encode the input question $q$ using BERT~\cite{devlin2019bert} and then decode the \texttt{aggregation}, \texttt{target}, \texttt{condition} separately as described in~\autoref{fig:baseline}. The \texttt{aggregation} slot can have the following values of ``argmax, argmin, argmax-date, argmin-date", the \texttt{target} and \texttt{condition} slots have their potential values based on the table field and its corresponding entries. Though we do not have the ground-truth annotation for these simple SQL queries, we can use heuristics to infer them from the denotation. We use the synthesized question-SQL pairs to train the parser model.

\subsection{Passage-Only Model}
In this setting, we design a model that only uses the hyperlinked passages from the given table to find the answer. Our model is based on DrQA~\cite{chen2017reading}, which first uses an ensemble of several retrievers to retrieve related documents and then concatenate several documents together to do reading comprehension with the state-of-the-art BERT model~\cite{devlin2019bert}. The basic architecture is depicted in~\autoref{fig:baseline}, where we use the retriever to retrieve the top-5 passages from the pool and then concatenate them as a document for the MRC model, and the maximum length of the concatenated document is set to 512.  

\subsection{\model}
\begin{figure}[!t]
    \centering
    \includegraphics[width=1.0\linewidth]{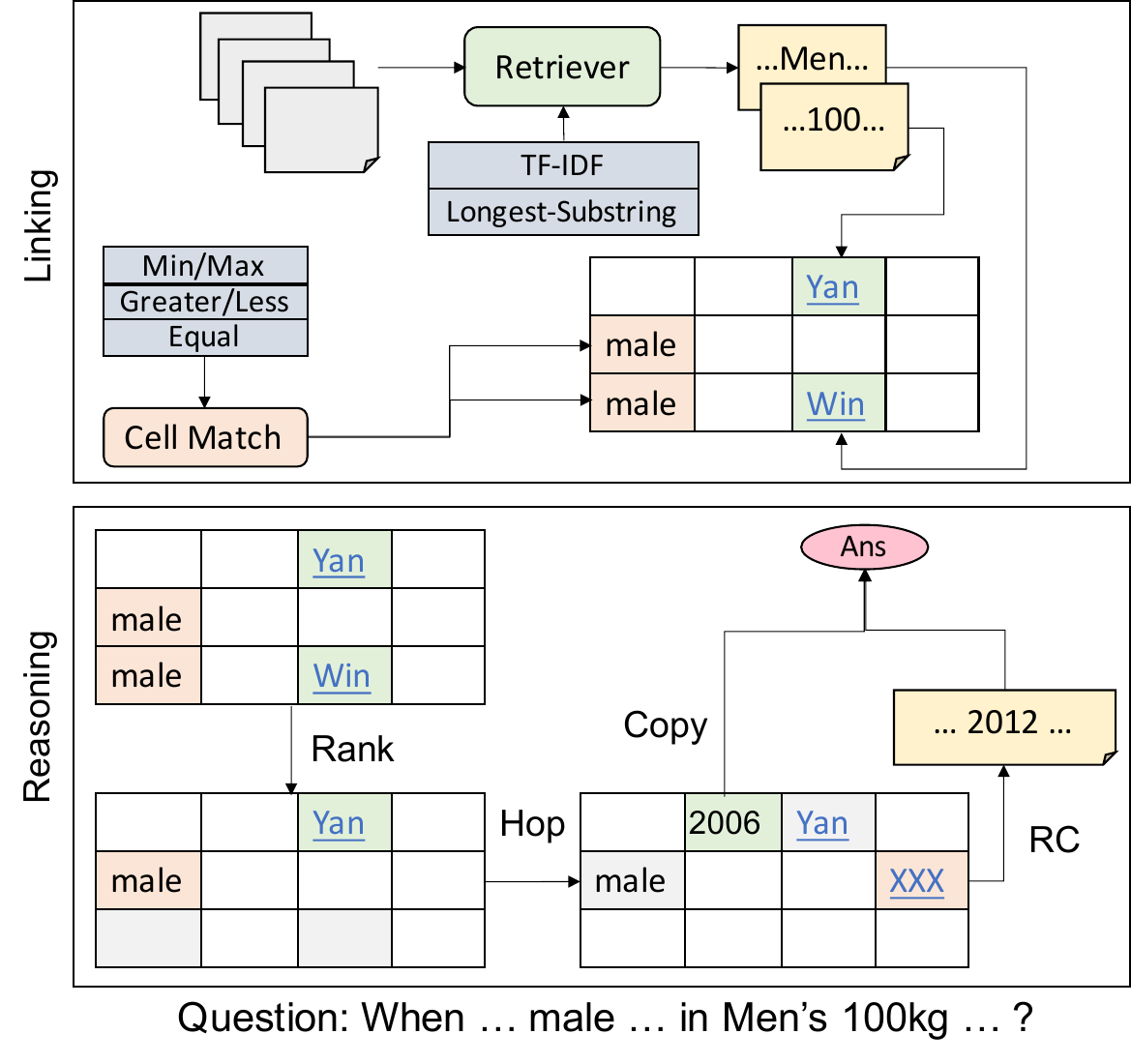}
    \vspace{-2ex}
    \caption{Illustration of the proposed model to perform multi-hop reasoning over table and passage.}
    \label{fig:mixer}
    \vspace{-2ex}
\end{figure}

In order to cope with heterogeneous information, we propose a novel architecture called \model. We divide the model into two phases as depicted in~\autoref{fig:mixer} and describe them separately below:
\paragraph{Linking}
This phase is aimed to link questions to their related cells from two sources: \\
\noindent - Cell Matching: it aims to link cells explicitly mentioned by the question. The linking consists of three criteria, 1) the cell's value is explicitly mentioned, 2) the cell's value is greater/less than the mentioned value in question, 3) the cell's value is maximum/minimum over the whole column if the question involves superlative words.\\
\noindent - Passage Retriever: it aims to link cells implicitly mentioned by the question through its hyperlinked passage. The linking model consists of a TD-IDF retriever with 2-3 gram lexicon and a longest-substring retriever, this ensemble retriever calculates the distances with all the passages in the pool and highlight the ones with cosine distance lower than a threshold $\tau$. The retrieved passages are mapped back to the linked cell in the table.
\begin{figure}[!ht]
    \centering
    \includegraphics[width=1.0\linewidth]{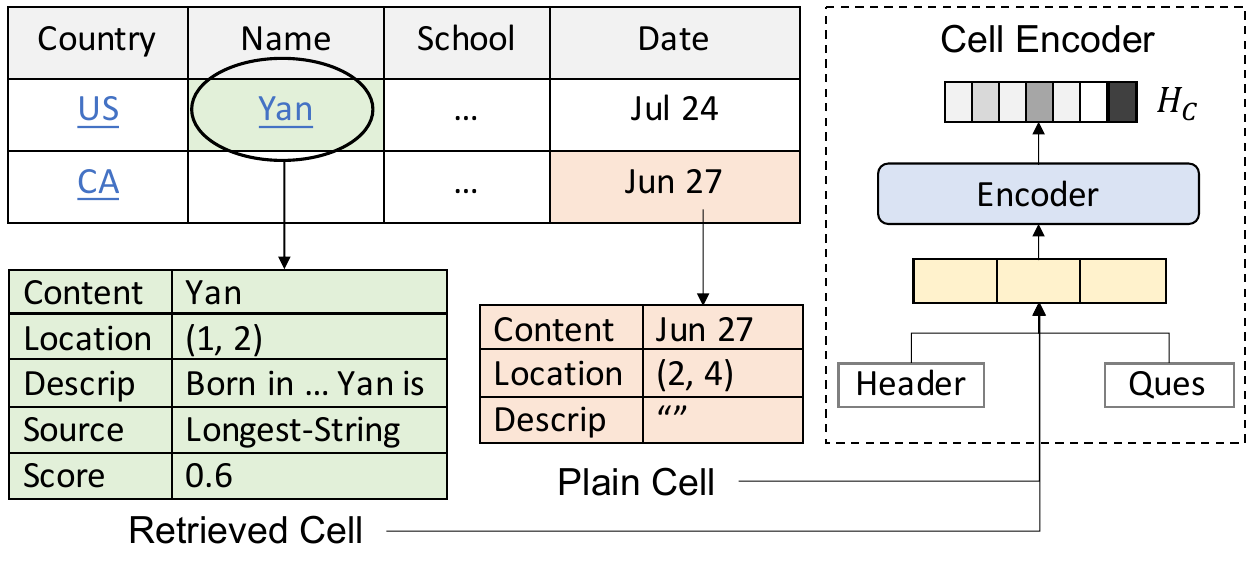}
    \caption{Illustration of cell encoder of retrieved (green) and plain cells (orange).}
    \label{fig:cell}
    \vspace{-2ex}
\end{figure}

We call the set of cells from these two sources as ``retrieved cells" denotes by $\mathcal{C}$. Each retrieved cell $c$ is encoded by 5-element tuple \texttt{(content, location, description, source, score)}. \texttt{Content} represents the string representation in the table, \texttt{Content} refers to the absolute row and column index in the table, \texttt{description} refers to the evidence sentence in the hyperlinked passage, which gives highest similarity score to question, \texttt{source} denotes where the entry comes from (e.g. equal/argmax/passage/etc), \texttt{score} denotes the score of linked score normalized to [0, 1].

\paragraph{Reasoning}
This phase is aimed to model the multi-hop reasoning in the table and passage, we specifically break down the whole process into three stages, namely the ranking stage $p_{f}(c | q, \mathcal{C})$, hoping stage $p_{h}(c' | q, c)$, and the reading comprehension stage $p_{r}(a | P, q)$. These three stages are modeled with three different neural networks. We first design a cell encoding scheme to encode each cell in the table as depicted in~\autoref{fig:cell}: 1) for ``retrieved cells", it contains information for retrieval source and score, 2) for ``plain cells" (not retrieved), we set the information in source and score to empty. We concatenate them with their table field and question, and then fed into a encoder module (BERT) to obtain its vector representation $H_c$.

\vspace{2ex}
\noindent 1) Ranking model: As the ``retriever cells`` contain many noises, we leverage a ranker model to predict the ``correct" linked cells for the next stage. Specifically, this model takes each cell $c$ along with its neighboring $\mathcal{N}_c$ (cells in the same row) and feed them all into the cell encoder to obtain their representations $\{H_c\}$. The representations are aggregated and  further fed to a feed-forward neural network to obtain a score $s_c$, which is normalized over the whole set of linked cell $\mathcal{C}$ as follows:
\begin{equation}
\small
p_{f}(c|q, \mathcal{C}) = \frac{exp(s_c)}{\sum_{c' \in \mathcal{C}} exp(s_{c'})}
\end{equation}
\noindent 2) Hop model: this model takes the predicted cell from the previous stage and then decide which neighboring cell or itself to hop to. Specifically, we represent each hop pair $(c \rightarrow c')$ using their concatenated representation $H_{c, c'}=[H_c, H_{c'}]$. The representation is fed to a feed-forward neural network to obtain a hop score $s_{c, c'}$, which is normalized over all the possible end cells as follows:
\begin{equation}
\small
p_{f}(c'|q, c) = \frac{exp(s_{c, c'})}{\sum_{c'' \in \mathcal{N}_c \cup c} exp(s_{c, c''})}
\end{equation}
\noindent 3) RC model: this model finally takes the hopped cell $c$ from last stage and find answer from it. If the cell is not hyperlinked, the RC model will simply output its plain text as the answer, otherwise, the plain text of the cell is prepended to the linked passage $P(c)$ for reading comprehension. The prepended passage $P$ and the question are given as the input to the question answering model to predict the score of answer's start and end index as $g_s(P, q, index)$ and $g_e(P, q, index)$, which are normalized over the whole passage $|P|$ to calculate the likelihood $p_{r}(a|P, q)$ as follows:
\begin{align*}
\small
\begin{split}
    p_{r}(a|P, q) = \frac{exp(g_s(P, q, a_s))}{\sum_{i \in |P|} exp(g_s(P, q, i))} \frac{g_s(P, q, a_e)}{\sum_{i \in |P|} g_e(P, q, i)}
\end{split}
\end{align*}
where $a_s$ is the start index of answer $a$ and $a_e$ is the end index of answer $a$.

By breaking the reasoning process into three stages, we manage to cover the Type-I/II/III/VI questions well. For example, the Type-III question first uses the ranking model to select the most likely cell from retrievers, and then use the hop model to jump to neighboring hyperlinked cell, finally use the RC model to extract the answer.

\paragraph{Training \& Inference}
The three-stage decomposition breaks the question answering likelihood $p(a|q, T)$ into the following marginal probability: 
\begin{align*}
\small
\begin{split}
    \sum_{c \in \mathcal{C}}p_f(c|q, \mathcal{C}) \sum_{c' \in \mathcal{N}_c; a \in P(c')} p_f(c'|c, q) p_r(a|P(c'), q)
\end{split}
\end{align*}
where the marginalization is over all the linked cells $c$, and all the neighboring cell with answer $a$ in its plain text or linked passages. However, directly maximizing the marginal likelihood is unnecessarily complicated as the marginalization leads to huge computation cost. Therefore, we propose to train the three models independently and then combine them to do inference.

By using the source location of answers, we are able to 1) infer which cells $c$ in the retrieved set $\mathcal{C}$ are valid, which can be applied to train the ranking model, 2) infer which cell it hops to get the answer, which we can be applied to train the hop model. Though the synthesized reasoning paths are somewhat noisy, it is still enough to be used for training the separate models in a weakly supervised manner. For the RC model, we use the passages containing the ground-truth answer to train it. The independent training avoids the marginalization computation to greatly decrease the computation and time cost. During inference, we apply these three models sequentially to get the answer. Specifically, we use greedy search at first two steps to remain only the highest probably cell and finally extract the answer using the RC model. 

\begin{table*}[!th]
\small
\begin{tabular}{l|ccc|ccc}
\hline
\multicolumn{1}{l|}{\multirow{3}{*}{Model}} & \multicolumn{3}{c|}{Dev} & \multicolumn{3}{c}{Test}    \\ 
\cline{2-7} 
\multicolumn{1}{l|}{}                      & \multicolumn{1}{c}{In-Table} & \multicolumn{1}{c}{In-Passage} & \multicolumn{1}{c|}{Total} & \multicolumn{1}{c}{In-Table} & \multicolumn{1}{c}{In-Passage} & \multicolumn{1}{c}{Total}   \\
\cline{2-7}
\multicolumn{1}{l|}{}                       & EM/F1   & EM/F1 & EM/F1    & EM/F1       & EM/F1 & EM/F1   \\ 
\hline
Table-Only                                  & 14.7/19.1  & 2.4/4.5          & 8.4/12.1   & 14.2/18.8  &  2.6/4.7    & 8.3/11.7 \\
Passage-Only                                & 9.2/13.5   & 26.1/32.4        & 19.5/25.1  &  8.9/13.8  &   25.5/32.0 & 19.1/25.0 \\ 
\hline
\model (BERT-base-uncased, $\tau$=0.7)      & 51.2/58.6  & 39.6/46.4   & 42.9/50.0     &  50.9/58.6   & 37.4/45.7  & 41.8/49.5      \\
\model (BERT-base-uncased, $\tau$=0.8)      & 51.3/58.4 & 40.1/47.6 & 43.5/50.6 & 51.7/59.1 & 37.8/46.0  & 42.2/49.9 \\
\model (BERT-base-uncased, $\tau$=0.9)      & 51.5/58.6 & 40.5/47.9    & 43.7/50.9  &  52.1/59.3  & 38.1/46.3  & 42.5/50.2   \\
\model (BERT-large-uncased, $\tau$=0.8)     & 54.3/61.4  &  39.1/45.7  &  \textbf{44.0}/\textbf{50.7}  & 56.2/63.3 & 37.5/44.4  & \textbf{43.8}/\textbf{50.6}   \\
\hline
\end{tabular}
\caption{Experimental results of different models, In-Table refers to the subset of questions which have their answers in the table, In-Passage refers to the subset of questions which have their answer in a certain passage.}
\label{tab:results}
\vspace{-2ex}
\end{table*}

\section{Experiments}
\subsection{Experimental Setting}
In the linking phase, we set the retrieval threshold $\tau$ to a specific value. All the passages having distance lower than $\tau$ will be retrieved and fed as input to the reasoning phase. If there is no passage that has been found with a distance lower than $\tau$, we will simply use the document with the lowest distance as the retrieval result. Increasing $\tau$ can increase the recall of correct passages, but also increase the difficulty of the filter model in the reasoning step. 

In the reasoning phase, we mainly utilize BERT~\cite{devlin2019bert} as our encoder for the cells and passages due to its strong semantic understanding. Specifically, we use four BERT variants provided by huggingface library\footnote{\url{https://github.com/huggingface/transformers}}, namely base-uncased, based-cased, large-uncased, and large-cased. We train the modules all for 3.0 epochs and save their checkpoint file at the end of each epoch. The filtering, hop, and RC models use AdamW~\cite{loshchilov2017decoupled} optimizer with learning rates of 2e-6, 5e-6, and 3e-5. We held out a small development set for model selection on the saved checkpoints and use the most performant ones in inference.

\subsection{Evaluation}
Following previous work~\cite{rajpurkar2016squad}, we use exact match (EM) and F1 as two evaluation metrics. F1 metric measures the average overlap between the prediction and ground-truth answers. We assess human performance on a held-out set from the test set containing 500 instances. To evaluate human performance, we distribute each question along with its table to crowd-workers and compare their answer with the ground-truth answer. We obtain an estimated accuracy of EM=88.2 and F1=93.5, which is higher than both SQuAD~\cite{rajpurkar2016squad} and HotpotQA~\cite{yang2018hotpotqa}. The higher accuracy is due to the In-Table questions (over 40\%), which have much lesser ambiguity than the text-span questions. 

\subsection{Experimental Results}
We demonstrate the experimental results for different models in~\autoref{tab:results}, where we list fine-grained accuracy over the questions with answers in the cell and passage separately.  The In-Table questions are remarkably simpler than In-Passage question because they do not the RC reasoning step; the overall accuracy is roughly 8-10\% higher than its counterpart. With the experimented model variants, the best accuracy is achieved with BERT-large-uncased as backend, which can beat the BERT-base-uncased by roughly 2\%. However, its performance is still far lagged behind human performance, leaving ample room for future research.

\paragraph{Heterogeneous Reasoning}
From~\autoref{tab:results}, we can clearly observe that using either Table-Only or Passage-Only model achieves a poor accuracy below 20\%. In contrast, the proposed \model can achieve up to 50\% EM increase by leveraging both structured and unstructured data during reasoning. It strongly supports the necessity to do heterogeneous reasoning in \dataset.
\paragraph{Retriever Threshold}
We also experiment with a different $\tau$ threshold. Having an aggressive retriever increases the recall of the mentioned cells, but it increases the burden for the ranking model. Having a passive retriever can guarantee the precision of predicted cells, but it also potentially miss evidence for the following reasoning phase. There exist trade-offs between these different modes. In~\autoref{tab:results}, we experiment with different $\tau$ during the retrieval stage and find that the model is rather stable, which means the model is quite insensitive regarding different threshold values.

\subsection{Error Analysis}
\begin{figure}[!ht]
    \centering
    \includegraphics[width=0.9\linewidth]{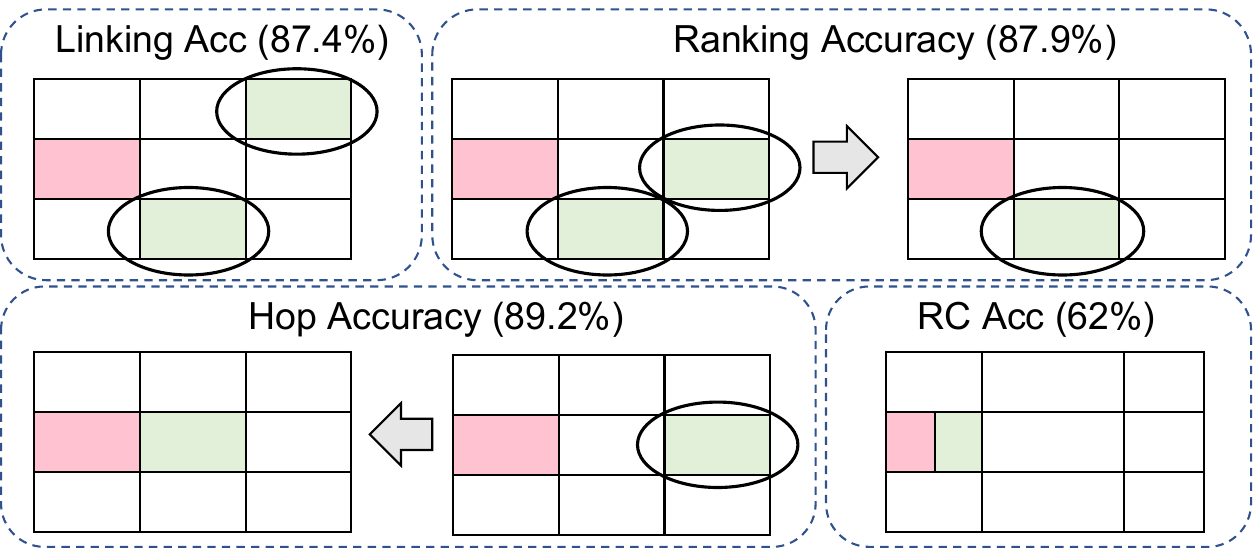}
    \caption{The error of \model is based on its stages. Pink cell means the answer cell; green means the model's prediction; circle means the current cell.}
    \label{fig:analysis}
    \vspace{-1ex}
\end{figure}

To analyze the cause of the errors in \model, we propose to break down into four types as~\autoref{fig:analysis}. Concretely, linking error is caused by the retriever/linker, which fails to retrieve the most relevant cell in the linking phase. In the reasoning phase: 1) ranking error is caused by the ranking model, which fails to assign a high score to the correct retrieved cell. 2) hop error occurs when the correctly ranked cell couldn't hop to the answer cell. 3) RC error refers to the hoped cell is correct, but the RC model fails to extract the correct text span from it.
We perform our analysis on the full dev set based on the bert-large-uncased model ($\tau$=0.8), as indicated in~\autoref{fig:analysis}, the errors are quite uniformly distributed into the four categories except the reading comprehension step is slightly more erroneous. Based on the step-wise error, we can compute its product as $87.4\% \times 87.9\% \times 89.2\% \times 61.9\% \approx 42\%$ and find that the result consistent well the overall accuracy, which demonstrates the necessity to perform each reasoning step correctly. Such error cascading makes the problem extremely difficult than the previous homogeneous question answering problems. 

By breaking down the reasoning into steps, \model layouts strong explainability about its rationale, but it also causes error propagation, i.e., the mistakes made in the earlier stage are non-reversible in the following stage. We believe future research on building an end-to-end reasoning model could alleviate such an error propagation problem between different stages in \model.   

\section{Related Work}
\paragraph{Text-Based QA} Since the surge of SQuAD~\cite{rajpurkar2016squad} dataset, there have been numerous efforts to tackle the machine reading comprehension problem. Different datasets like DrQA~\cite{chen2017reading}, TriviaQA~\cite{joshi2017triviaqa}, SearchQA~\cite{dunn2017searchqa} and DROP~\cite{dua2019drop} have been proposed. As the SQuAD~\cite{rajpurkar2016squad} questions that are relatively simple because they usually require no more than one sentence in the paragraph to answer. The following datasets further challenge the QA model's capability to handle different scenarios like open-domain, long context, multi-hop, discrete operations, etc. There has been a huge success in proving that the deep learning model can show strong competence in terms of understanding text-only evidence. Unlike these datasets, \dataset leverages structured information in the evidence form, where the existing models are not able to handle, which distinguishes it from the other datasets.

\paragraph{KB/Table-Based QA} Structured knowledge is known as unambiguous and compositional, which absorbs lots of attention to the QA system built on KB/Tables. There have been multiple datasets like WebQuestion~\cite{berant2013semantic}, ComplexWebQuestions~\cite{talmor2018web}, WebQuestionSP~\cite{yih2015semantic} on using FreeBase to answer natural questions. Besides KB, structured or semi-structured tables are also an interesting form. Different datasets like WikiTableQuestions~\cite{pasupat2015compositional}, WikiSQL~\cite{zhong2017seq2sql}, SPIDER~\cite{yu2018spider}, TabFact~\cite{2019TabFactA} have been proposed to build models which can interact with such structured information. However, both KB and tables are known to suffer from low coverage issues. Therefore, \dataset combine tables with text as complementary information to answer natural questions.  

\paragraph{Information Aggregation} There are some pioneering studies on designing hybrid question answering systems to aggregate heterogeneous information.  GRAFT~\cite{sun2018open} proposes to use the early fusion system and use heuristics to build a question-specific subgraph that contains sentences from corpus and entities, facts from KB. PullNet~\cite{sun2019pullnet} improves over GRAFT to use an integrated framework that dynamically learns to retrieve and reason over heterogeneous information to find the best answers. More recently, KAReader~\cite{xiong2019improving} proposes to reformulate the questions in latent space by reading retrieved text snippets under KB-incomplete cases. These models simulate a `fake' KB-incomplete scenario by masking out triples from KB. In contrast, the questions in \dataset are inherently hybrid in the sense that it requires both information forms to reason, which makes our testbed more realistic.

\section{Conclusion}
We present \dataset, which is collected as the first hybrid question answering dataset over both tabular and textual data. We release the data to facilitate the current research on using heterogeneous information to answer real-world questions. We design \model as a strong baseline and offer interesting insights about the model. We believe \dataset is an interesting yet challenging next-problem for the community to solve.

\section*{Acknowledgement}
The authors would like to thank the anonymous reviewers and area chairs for their thoughtful comments. We would like to thank Amazon AWS Machine Learning Research Award for sponsoring the computing resources.

\bibliography{emnlp2020}
\bibliographystyle{acl_natbib}

\end{document}